\documentclass[theme=lavender]{techreport}

\usepackage{amsmath,amsfonts,bm}

\def\eqref#1{equation~\ref{#1}}

\def\1{\bm{1}}

\def\vzero{{\bm{0}}}

\def\vb{{\bm{b}}}

\def\vg{{\bm{g}}}
\def\vh{{\bm{h}}}

\def\vk{{\bm{k}}}

\def\vm{{\bm{m}}}

\def\vq{{\bm{q}}}
\def\vr{{\bm{r}}}
\def\vs{{\bm{s}}}
\def\vt{{\bm{t}}}
\def\vu{{\bm{u}}}
\def\vv{{\bm{v}}}

\def\vx{{\bm{x}}}

\def\vepsilon{{\bm{\epsilon}}}
\def\valpha{{\bm{\alpha}}}

\def\mA{{\bm{A}}}

\def\mC{{\bm{C}}}

\def\mH{{\bm{H}}}
\def\mI{{\bm{I}}}

\def\mK{{\bm{K}}}

\def\mM{{\bm{M}}}

\def\mO{{\bm{O}}}

\def\mQ{{\bm{Q}}}
\def\mR{{\bm{R}}}

\def\mU{{\bm{U}}}
\def\mV{{\bm{V}}}
\def\mW{{\bm{W}}}
\def\mX{{\bm{X}}}
\def\mY{{\bm{Y}}}
\def\mZ{{\bm{Z}}}

\DeclareMathAlphabet{\mathsfit}{\encodingdefault}{\sfdefault}{m}{sl}
\SetMathAlphabet{\mathsfit}{bold}{\encodingdefault}{\sfdefault}{bx}{n}

\newcommand{\E}{\mathbb{E}}

\newcommand{\R}{\mathbb{R}}

\usepackage{tikz}
\usetikzlibrary{shapes.geometric, arrows.meta, positioning, fit, backgrounds, calc, decorations.pathreplacing}

\newcommand{\model}{\textsc{Meet}}

\title{Scalable Peptide Design via\\Memory-Efficient Equivariant Transformer}

\author[1,2]{Rui Jiao\textsuperscript{*}}
\author[1,2]{Xiangzhe Kong\textsuperscript{*}}
\author[2]{Yinjun Jia\textsuperscript{*}}
\author[2,3]{Yijia Zhang}
\author[4]{Ziyi Yang}
\author[1,2]{Yang Liu}
\author[2,3]{Jianzhu Ma\textsuperscript{\dag}}

\affiliation[1]{Department of Computer Science and Technology, Tsinghua University\\}
\affiliation[2]{Institute for AI Industry Research, Tsinghua University\\}
\affiliation[3]{Department of Electronic Engineering, Tsinghua University\\}
\affiliation[4]{Department of Chemistry, Tsinghua University}

\contribution[*]{Equal contribution}

\abstract{%
Target-specific peptide design requires sequence and structure co-design under full atom geometric constraints.
Latent generative frameworks offer an effective route for this problem by compressing fine grained atomic structures into block level latent representations and performing conditional generation in a compact latent space.
However, the scalability of such systems depends heavily on the geometric backbone used throughout their encoding, decoding, and denoising components.
We introduce \model{} (\textbf{M}emory \textbf{E}fficient \textbf{E}quivariant \textbf{T}ransformer), an E(3) equivariant backbone for scalable atomistic peptide modeling.
\model{} maintains coupled invariant scalar and equivariant vector feature streams, while reformulating geometric computation around memory efficient attention.
It initializes vector features through global coordinate aggregation, incorporates pairwise distances through augmented query and key dot products, and injects covalent bond information through sparse bond adaptation.
Integrated into a VAE and latent diffusion pipeline for full atom peptide generation, \model{} achieves linear memory scaling with atom count and improves generation quality over existing peptide design methods.
Experiments on large scale AFDB derived datasets further show that the proposed backbone supports systematic model and data scaling, leading to better binding affinity, physical validity, and sample diversity.%
}

\correspondence{\email{majianzhu@tsinghua.edu.cn}}
\codes{https://github.com/jiaor17/MEET}

\begin{document}

\maketitle

\section{Introduction}

Designing peptides that bind a specified protein pocket is a central problem in structure based drug discovery~\citep{wang2022therapeutic}.
The task is a form of sequence and structure co-design, where peptide sequence, conformation, and pocket binding geometry are coupled through physical interactions.
Accurate modeling therefore requires full atom geometric reasoning over side chain packing, hydrogen bonding, shape complementarity, and local steric compatibility.
This makes E(3) equivariant architectures~\citep{thomas2018tensor,satorras2021en,deng2021vectorneuronsgeneralframework,schutt2021equivariant} a natural choice, since they respect the symmetries of three dimensional space while operating directly on molecular geometry.

Latent generative frameworks~\citep{rombach2022high,kong2025unimomo} offer a practical way to balance full atom fidelity with generative efficiency.
In this formulation, a VAE compresses fine grained atomic structures into block level latent points, so that local atomic details are preserved through reconstruction rather than represented explicitly throughout the diffusion process.
Generation is then performed over a compact latent graph conditioned on the target pocket, with the decoder mapping the generated latent states back to full atom geometry.
As the geometric backbone is instantiated throughout the VAE and LDM components, its expressiveness and memory efficiency become central to the scalability of the overall generative system.

This places a strong demand on the backbone used for full atom peptide and pocket complexes.
Such complexes often contain hundreds to thousands of atoms.
Existing geometric backbones commonly introduce coordinate dependence either through a dense distance matrix~\citep{jiao2025equivariantpretrainedtransformerunified}, or through explicit local molecular graphs~\citep{schutt2018schnet,satorras2021en,schutt2021equivariant,tholke2022torchmdnet,liao2023equiformer}.
These mechanisms are effective, but they require storing or recomputing pairwise biases, neighbor lists, or edge features, which increases memory traffic and restricts the feasible model size and batch size.
The issue becomes more pronounced when scaling generative models, where improvements in sample quality often require both larger backbones and larger training sets.

To address this backbone level bottleneck, we introduce \model{} (\textbf{M}emory \textbf{E}fficient \textbf{E}quivariant \textbf{T}ransformer), an E(3) equivariant Transformer backbone for scalable atomistic peptide modeling.
\model{} keeps coupled scalar and vector feature streams, where the scalar stream is rotation invariant and the vector stream is rotation equivariant.
Its geometric operations are written in forms that are compatible with memory efficient attention kernels such as FlashAttention~\citep{dao2022flashattention}.
In particular, \model{} initializes vector features through global coordinate aggregation, folds pairwise distance information into augmented query and key dot products, and injects covalent bond information through a sparse bond adapter.
Together these design choices avoid materializing quadratic activation tensors when memory efficient attention is used, while preserving E(3) equivariance of the full backbone.

\begin{figure}[t]
  \centering
    \includegraphics[width=\linewidth]{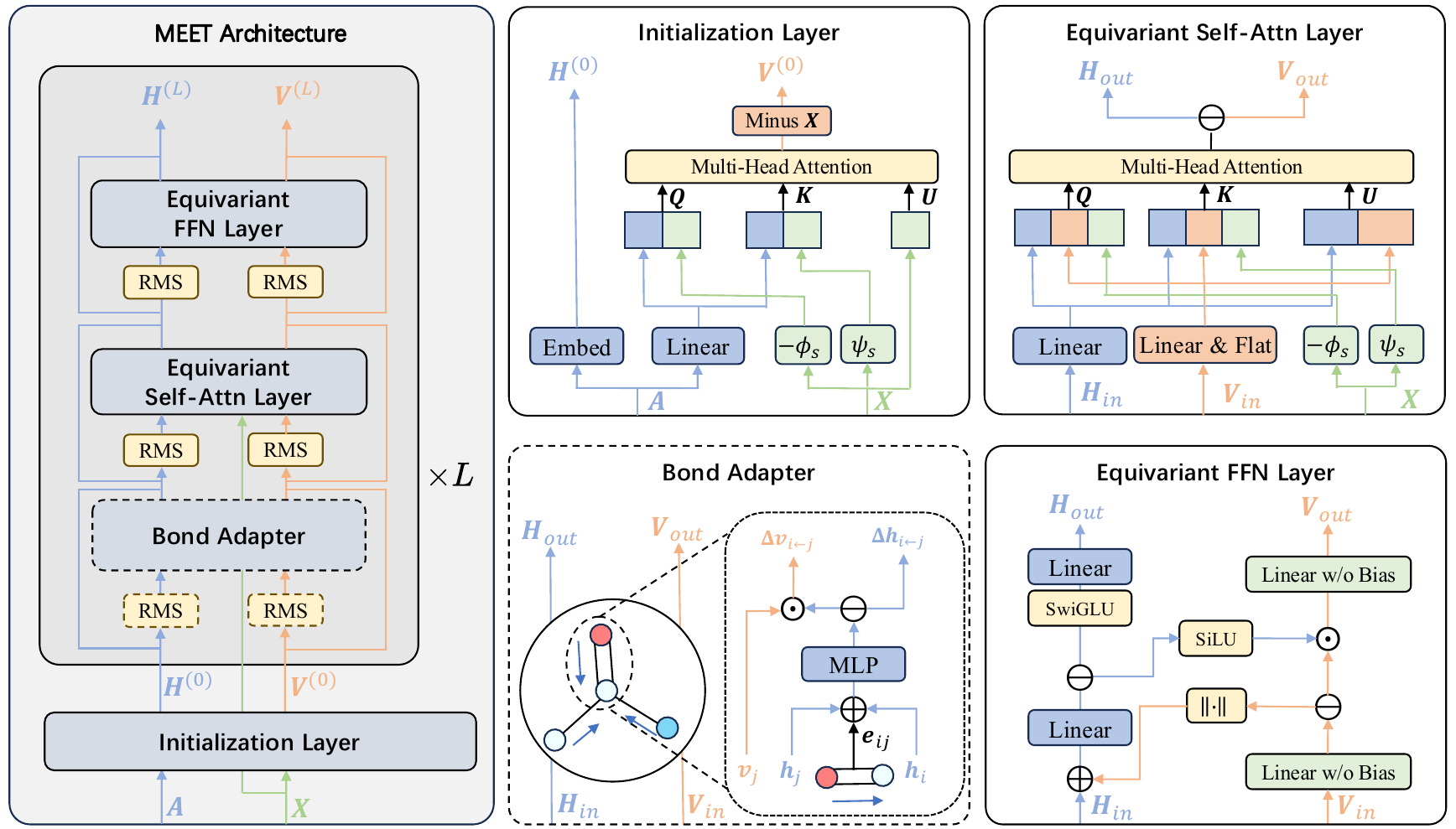}
\caption{
Overview of the \model{} architecture.
\model{} processes atom coordinates $\mX$ together with scalar and vector feature streams $\mH$ and $\mV$.
The initialization layer uses distance aware attention to produce $\mH^{(0)}$ and $\mV^{(0)}$.
Each repeated block contains a bond adapter, an equivariant self attention layer, and an equivariant feed forward layer, producing final features $\mH^{(L)}$ and $\mV^{(L)}$.
In self attention, scalar and vector inputs are projected into $\mQ$, $\mK$, and $\mU$, with distance encodings $-\phi_s$ and $\psi_s$ appended to $\mQ$ and $\mK$.
The bond adapter injects edge attributes $e_{ij}$ through sparse messages $\Delta \vh_{i\leftarrow j}$ and $\Delta \vv_{i\leftarrow j}$.
}
\label{fig:overview}
\end{figure}

We evaluate this backbone inside a two stage latent generative framework inspired by UniMoMo~\citep{kong2025unimomo}.
Within this framework, \model{} serves as the geometric backbone for the encoder, block type decoder, structure decoder, and LDM denoiser.
This setting lets us test whether a more memory efficient equivariant backbone improves the complete generation system.
Figure~\ref{fig:overview} summarizes the architecture and the main geometric modules.
Our scaling experiments focus on the LDM denoiser, whose capacity directly affects the quality of generated peptides in our framework.

Together, the architecture and generative evaluation lead to three concrete contributions.

\textbf{Efficient architecture.}
\model{} replaces dense distance biases with query and key augmentation, replaces local-graph vector initialization with global attention aggregation, and uses sparse bond adaptation for chemical adjacency.
These choices give linear peak activation memory in the number of atoms for a fixed model size when paired with memory efficient attention kernels.

\textbf{Large scale datasets.}
Starting from $8.64$ million AFDB~\citep{varadi2022alphafold} domains, we construct approximately $100$ million candidate segments and sample $100\mathrm{K}$ and $1.2\mathrm{M}$ training structures using sliding window enumeration, structure quality filtering, interface screening, and sequence overlap clustering.

\textbf{Systematic scaling.}
On the $100\mathrm{K}$ benchmark, \model{} improves over PepGLAD~\citep{kong2024pepglad}, PepFlow~\citep{li2024pepflow}, UniMoMo~\citep{kong2025unimomo}, and DiffPepBuilder~\citep{wang2024diffpepbuilder} on binding free energy and physical validity.
Scaling the latent denoiser across four DiT-style~\citep{peebles2023scalable} model sizes on $1.2\mathrm{M}$ datasets further improves generation quality and restores sample diversity.

\section{Methods}

Our peptide generator follows a two-stage latent generative framework inspired by UniMoMo~\citep{kong2025unimomo}. A VAE first encodes each full-atom peptide--pocket complex into block-level latent variables, a conditional latent diffusion model generates peptide latents from the target-pocket context, and the VAE decoder maps the generated latents back to peptide sequence and full-atom geometry. \model{} serves as the geometric backbone in the VAE encoder, sequence decoder, structure decoder, and latent denoiser. The complete training objectives and sampling procedure are provided in Appendix~\ref{app:latent-framework}.

In the following section, we first present an overview of the \model{} architecture in \S\ref{sec:arch}. We then describe the distance-aware attention mechanism (\S\ref{sec:dist-attn}) that is shared across the backbone, followed by the four main modules shown in Figure~\ref{fig:overview}, including feature initialization (\S\ref{sec:init}), equivariant self-attention (\S\ref{sec:attn}), equivariant feed-forward layer (\S\ref{sec:ffn}), and bond adapter (\S\ref{sec:edge}). Finally, we analyze the space complexity in \S\ref{sec:complexity}.

\subsection{Architecture Overview}
\label{sec:arch}

\model{} is a memory-efficient equivariant Transformer that maintains two coupled feature streams, scalar and vector. We use the same notation as Figure~\ref{fig:overview}, where $\mH$ denotes the scalar stream, $\mV$ denotes the vector stream, and $\mX$ denotes atom coordinates. For a molecule with $N$ atoms, let $\mX \in \R^{N \times 3}$ have rows $\vx_i \in \R^3$. We represent per-atom features as a pair $(\mH, \mV)$, where $\mH \in \R^{N \times d}$ are scalar (rotation-invariant) features and $\mV \in \R^{N \times 3 \times d}$ are vector (rotation-equivariant) features, following the common scalar-vector decomposition used in equivariant neural networks~\citep{deng2021vectorneuronsgeneralframework}. Under a rigid transformation $(\mR, \vt)$ with $\mR \in \mathrm{SO}(3)$ and $\vt \in \R^3$, features transform as $\mH \mapsto \mH$ and $\mV \mapsto \mR\,\mV$. All modules of \model{} are therefore E(3)-equivariant under the full Euclidean group.

Given input atom attributes $\mA$ and coordinates $\mX$, the backbone first constructs an initial equivariant state via an initialization operator $\mathcal{I}$,
\begin{equation}
  (\mH^{(0)}, \mV^{(0)}) \;=\; \mathcal{I}\bigl(\mA, \mX\bigr),
\end{equation}
and then applies $L$ Transformer blocks to produce $(\mH^{(L)}, \mV^{(L)})$. Each block consists of a self-attention layer $\mathcal{A}$ and a vector--scalar mixing feed-forward layer $\mathcal{F}$, optionally preceded by a bond adapter $\mathcal{B}$ when a bond edge set $\mathcal{E}$ is provided.
\begin{align}
  (\bar{\mH}^{(\ell)}, \bar{\mV}^{(\ell)}) &\;=\; (\mH^{(\ell)}, \mV^{(\ell)}) \;+\; \mathcal{B}^{(\ell)}\!\bigl(\mathrm{Norm}(\mH^{(\ell)}, \mV^{(\ell)}),\, \mathcal{E}\bigr), \\
  (\tilde{\mH}^{(\ell)}, \tilde{\mV}^{(\ell)}) &\;=\; (\bar{\mH}^{(\ell)}, \bar{\mV}^{(\ell)}) \;+\; \mathcal{A}^{(\ell)}\!\bigl(\mathrm{Norm}(\bar{\mH}^{(\ell)}, \bar{\mV}^{(\ell)}),\, \mX\bigr), \\
  (\mH^{(\ell+1)}, \mV^{(\ell+1)}) &\;=\; (\tilde{\mH}^{(\ell)}, \tilde{\mV}^{(\ell)}) \;+\; \mathcal{F}^{(\ell)}\!\bigl(\mathrm{Norm}(\tilde{\mH}^{(\ell)}, \tilde{\mV}^{(\ell)})\bigr).
\end{align}
Each layer is wrapped with pre-normalization and a residual connection, following standard Transformer practice. In module-level descriptions below, $(\mH_{\mathrm{in}}, \mV_{\mathrm{in}})$ and $(\mH_{\mathrm{out}}, \mV_{\mathrm{out}})$ refer to the streams entering and leaving the corresponding panel in Figure~\ref{fig:overview}. The bond adapter can be applied in every block, only in the first block, or disabled entirely depending on bond availability.

\paragraph{Normalization.}
For the scalar stream we use RMSNorm~\citep{zhang2019root}. For the vector stream we use a rotation-invariant analogue that normalizes by the combined root-mean-square over the spatial and channel axes of each atom's vector features,
\begin{equation}
  \mathrm{RMSNorm}_V(\mV_i) \;=\; \frac{\mV_i}{\sqrt{\dfrac{1}{3d}\sum_{k,c} \mV_{i,k,c}^{2} + \epsilon}}.
\end{equation}
The scale factor in the denominator is a rotation invariant, so the output transforms as $\mR\,\mV_i$ under a rotation and remains equivariant.

\subsection{Distance-aware Attention Mechanism}
\label{sec:dist-attn}

A component shared by the initialization layer $\mathcal{I}$ and the self-attention layers $\mathcal{A}$ is a distance-aware query--key augmentation. For head $h$, the attention logit includes an invariant distance penalty,
\begin{equation}
  A_{ij}^{(h)} \;=\; \rho_h\Bigl(\langle \vq_i^{(h)}, \vk_j^{(h)} \rangle \;-\; s_h^{2}\,\lVert \vx_i - \vx_j\rVert_2^{2}\Bigr),
  \label{eq:dist-logit}
\end{equation}
where $\vq_i^{(h)}$ and $\vk_j^{(h)}$ are the module-specific query and key features, $\rho_h$ is the attention scale, and $s_h$ is a learnable distance scale.

According to FlashBias~\citep{wu2026flashbias}, rather than storing the distance term as an $N\times N$ bias tensor, we absorb it into the query--key dot product. For coordinates $\vx_i$ with components $(x_{1,i}, x_{2,i}, x_{3,i})$, define
\begin{equation}
  \phi_{s_h}(\vx_i) \;=\; s_h \bigoplus_{k=1}^{3}\bigl[x_{k,i}^{2},\; 1,\; -2 x_{k,i}\bigr],
  \qquad
  \psi_{s_h}(\vx_j) \;=\; s_h \bigoplus_{k=1}^{3}\bigl[1,\; x_{k,j}^{2},\; x_{k,j}\bigr].
  \label{eq:dist-enc}
\end{equation}
Since $\langle \phi_{s_h}(\vx_i), \psi_{s_h}(\vx_j)\rangle = s_h^{2}\lVert \vx_i - \vx_j\rVert_2^{2}$, concatenating $-\phi_s$ to the query and $\psi_s$ to the key reproduces Equation~\eqref{eq:dist-logit},
\begin{equation}
  \mQ_i^{(h)} \;=\; \mathrm{concat}\bigl(\vq_i^{(h)},\; -\phi_{s_h}(\vx_i)\bigr),
  \qquad
  \mK_j^{(h)} \;=\; \mathrm{concat}\bigl(\vk_j^{(h)},\; \psi_{s_h}(\vx_j)\bigr),
  \qquad
  A_{ij}^{(h)} \;=\; \rho_h\langle \mQ_i^{(h)},\, \mK_j^{(h)}\rangle.
  \label{eq:dist-qkaug}
\end{equation}
The augmentation adds only $9$ entries per head and keeps the attention computation compatible with fused kernels without an external distance-bias tensor. The resulting softmax weights
\begin{equation}
  \alpha_{ij}^{(h)} \;=\; \mathrm{softmax}_j\bigl(A_{ij}^{(h)}\bigr)
\end{equation}
depend on coordinates only through pairwise distances and are therefore E(3)-invariant.

\subsection{Feature Initialization}
\label{sec:init}

The initialization layer $\mathcal{I}$ prepares the initial scalar and vector streams from atom attributes and coordinates. The scalar stream is obtained by embedding the input atom attributes, denoted as $\mH^{(0)}=\mathrm{Embed}(\mA)$. The main role of $\mathcal{I}$ is therefore to construct a non-trivial vector stream $\mV^{(0)}$ from geometry. This step is necessary because every subsequent operation on the vector stream is either a bias-free linear projection along the channel axis or a linear combination of equivariant vectors. If $\mV$ were initialized as zero, the vector branch would remain uninformative and could not carry directional information.

Earlier equivariant backbones typically obtain directional states by aggregating relative positions over explicit local molecular graphs~\citep{schutt2018schnet,satorras2021en,schutt2021equivariant,tholke2022torchmdnet,liao2023equiformer}. Such designs require constructing neighbor edges and storing edge features, which increases memory traffic for large full-atom complexes. We instead initialize vector features through one round of distance-aware multi-head attention over the atoms in a complex, keeping the initialization compatible with memory-efficient attention kernels without constructing an explicit local molecular graph.

Concretely, queries and keys are computed from $\mH^{(0)}$ and augmented with the distance encodings in Section~\ref{sec:dist-attn}, yielding invariant per-head attention weights $\alpha_{ij}^{(h)}$. We collect these weights into a per-head attention matrix $\valpha^{(h)} \in \R^{N \times N}$. Unlike a standard attention layer, the value in $\mathcal{I}$ is the coordinate matrix $\mX \in \R^{N \times 3}$ itself, shared across heads. For each head $h$, the attended coordinate is centered at the query atom,
\begin{equation}
  \mV^{(h)} \;=\; \valpha^{(h)} \mX - \mX \;=\; \bigl(\valpha^{(h)} - \mI\bigr)\mX \;\in\; \R^{N \times 3}.
\end{equation}
Equivalently, the vector at atom $i$ is
\begin{equation}
  \vv_i^{(h)}
  \;=\;
  \sum_j \alpha_{ij}^{(h)}(\vx_j-\vx_i).
\end{equation}
The centering by $\vx_i$ removes dependence on the absolute coordinate frame and ensures that only relative displacements enter the vector stream.

The $H$ per-head displacement fields are then stacked along a new channel axis and projected to the model dimension with a bias-free linear map,
\begin{equation}
  \mV^{(0)} \;=\; \mathrm{stack}_h\bigl(\mV^{(h)}\bigr)\,\mW_{\text{out}}^{\mathcal{I}} \;\in\; \R^{N \times 3 \times d},
\end{equation}
where $\mathrm{stack}_h$ forms a tensor in $\R^{N \times 3 \times H}$ before the projection mixes the head axis into $d$ vector channels while leaving the spatial axis untouched.
Because $\alpha_{ij}^{(h)}$ depends only on pairwise distances, it is invariant to rigid transformations. The centered displacement $\sum_j \alpha_{ij}^{(h)}(\vx_j-\vx_i)$ is invariant to translation and transforms as a 3D vector under rotation. The bias-free output projection only mixes channels, so $\mV^{(0)}$ is equivariant.

\subsection{Equivariant Self-Attention}
\label{sec:attn}

Each Transformer block applies an equivariant self-attention layer $\mathcal{A}$ that updates both streams in a single multi-head attention pass, mapping $(\mH_{\mathrm{in}}, \mV_{\mathrm{in}})$ to $(\mH_{\mathrm{out}}, \mV_{\mathrm{out}})$ as in Figure~\ref{fig:overview}.

\paragraph{Q, K, and U preparation.}
The scalar and vector streams are projected independently. An unconstrained linear layer maps $\mH_{\mathrm{in}}$ to $(\mH_Q, \mH_K, \mH_U)$, while a bias-free linear layer followed by flattening on the channel axis of $\mV_{\mathrm{in}}$ yields $(\mV_Q, \mV_K, \mV_U)$. The vector queries and keys are normalized via $\mathrm{RMSNorm}_V$ and their spatial axis is flattened into the channel axis. The per-head query and key features entering the distance-aware mechanism of Section~\ref{sec:dist-attn} are then
\begin{equation}
  \vq_i^{(h)} = \mathrm{concat}\bigl(\mH_{Q,i}^{(h)},\; \mathrm{flat}(\mV_{Q,i}^{(h)})\bigr),
  \qquad
  \vk_j^{(h)} = \mathrm{concat}\bigl(\mH_{K,j}^{(h)},\; \mathrm{flat}(\mV_{K,j}^{(h)})\bigr),
  \label{eq:attn-qk}
\end{equation}
which are augmented with the distance encoding vectors $-\phi_s$ and $\psi_s$ (Equation~\eqref{eq:dist-qkaug}) to form $\mQ_i^{(h)}$ and $\mK_j^{(h)}$ as before. The value $\mU_j^{(h)}$ concatenates only the scalar and flattened-vector parts, without any distance encoding.
\begin{equation}
  \mU_j^{(h)} = \mathrm{concat}\bigl(\mH_{U,j}^{(h)},\; \mathrm{flat}(\mV_{U,j}^{(h)})\bigr).
  \label{eq:attn-v}
\end{equation}
The dot product $\langle \mQ_i^{(h)}, \mK_j^{(h)}\rangle$ is E(3)-invariant because the scalar parts are trivially invariant, the flattened vector parts contribute a sum of 3D inner products $\sum_c \langle \mV_{Q,i,:,c}, \mV_{K,j,:,c}\rangle$ that are invariant by orthogonality of $\mR$, and the distance encoding contributes $-s_h^{2}\lVert \vx_i - \vx_j\rVert_2^{2}$.

\paragraph{Aggregation and output projection.}
The invariant attention weights $\alpha_{ij}^{(h)}$ are applied to $\mU^{(h)}$, and the result is split into scalar and vector halves.
\begin{equation}
  \mO_i^{(h)} = \sum_j \alpha_{ij}^{(h)}\, \mU_j^{(h)},
  \qquad
  \mO_i^{(h)} \to \bigl(\mO_{H,i}^{(h)},\; \mO_{V,i}^{(h)}\bigr).
\end{equation}
The per-head outputs are concatenated across heads and projected back to the model dimension by two independent output projections.
\begin{equation}
  \mH_{\mathrm{out},i} = \mathrm{concat}_h\!\bigl(\mO_{H,i}^{(h)}\bigr)\,\mW_{\text{out}}^{H},
  \qquad
  \mV_{\mathrm{out},i} = \mathrm{concat}_h\!\bigl(\mO_{V,i}^{(h)}\bigr)\,\mW_{\text{out}}^{V},
\end{equation}
where $\mW_{\text{out}}^{H}$ is unconstrained and $\mW_{\text{out}}^{V}$ is bias-free and acts on the channel axis only. The scalar branch manipulates invariant quantities throughout, and the vector branch applies only bias-free channel mixing to equivariant tensors, so E(3) equivariance is preserved.

\subsection{Equivariant Feed-Forward Layer}
\label{sec:ffn}

Each attention layer is followed by a feed-forward layer $\mathcal{F}$ that couples the scalar and vector streams while preserving E(3) equivariance. The layer maps $(\mH_{\mathrm{in}}, \mV_{\mathrm{in}})$ to $(\mH_{\mathrm{out}}, \mV_{\mathrm{out}})$ as in Figure~\ref{fig:overview}. The vector features are first projected through a bias-free linear layer and split into a scalar-summary part and a hidden part.
\begin{equation}
  (\mV_{\mathrm{in}}^{(1)},\; \mV_{\mathrm{in}}^{(2)}) \;=\; \mathrm{split}\bigl(\mV_{\mathrm{in}}\,\mW_V^{\text{in}}\bigr), \qquad \mV_{\mathrm{in}}^{(1)} \in \R^{N \times 3 \times d},\; \mV_{\mathrm{in}}^{(2)} \in \R^{N \times 3 \times d_{\text{ff}}}.
\end{equation}
An invariant summary $\mY_i = [\lVert \mV_{\mathrm{in},i,:,1}^{(1)}\rVert_2, \ldots, \lVert \mV_{\mathrm{in},i,:,d}^{(1)}\rVert_2] \in \R^{d}$ is computed by taking channel-wise norms. The scalar branch uses the SwiGLU nonlinearity~\citep{shazeer2020glu}, and the scalar and vector streams are then updated as
\begin{align}
  (\vh^{(1)},\; \vh^{(2)}) &\;=\; \mathrm{split}\bigl([\mH_{\mathrm{in}}, \mY]\,\mW_H^{\text{in}} + \vb_H^{\text{in}}\bigr), \\
  \vh^{(2)} &\;\leftarrow\; \mathrm{SwiGLU}(\vh^{(2)}), \qquad \mV_{\mathrm{in}}^{(2)} \;\leftarrow\; \mathrm{SiLU}(\vh^{(1)}) \odot \mV_{\mathrm{in}}^{(2)}, \\
  \mH_{\mathrm{out}} &\;=\; \vh^{(2)}\,\mW_H^{\text{out}} + \vb_H^{\text{out}}, \qquad \mV_{\mathrm{out}} \;=\; \mV_{\mathrm{in}}^{(2)}\,\mW_V^{\text{out}}.
\end{align}

\subsection{Bond Adapter}
\label{sec:edge}

Geometric proximity alone does not fully determine local interactions. Chemical adjacency, covalent bonds, and other structured edge signals encode strong constraints that a dense attention mechanism captures only indirectly. Unlike the dense pairwise distance bias, bond adjacency is sparse and low-degree, so treating it as a dense attention bias would be wasteful. We instead inject bond information through a sparse message-passing adapter $\mathcal{B}$ whose cost scales linearly with the number of edges.

Let $\mathcal{E}$ denote the bond edge set and let $e_{ij} \in \R^{d_e}$ be an edge attribute for $(i,j) \in \mathcal{E}$. For each edge, we use the notation in Figure~\ref{fig:overview}, where $\vh_i$ and $\vh_j$ are the scalar features of the target and source atoms, and $\vv_j$ is the source atom's vector feature. We concatenate the scalar endpoint features with the edge attribute and pass the result through a small MLP $f_\theta$.
\begin{equation}
  \vm_{ij} \;=\; f_\theta\bigl([\vh_i,\, \vh_j,\, e_{ij}]\bigr) \;\in\; \R^{2d}.
\end{equation}
The output $\vm_{ij}$ is split into a scalar message $\Delta \vh_{i\leftarrow j} \in \R^{d}$ and a gating coefficient $\vg_{i\leftarrow j} \in \R^{d}$. The scalar messages are aggregated by mean-pooling over incoming edges, while the vector message $\Delta \vv_{i\leftarrow j}$ is constructed by gating the source vector feature $\vv_j$ with the scalar coefficient $\vg_{i\leftarrow j}$.
\begin{equation}
  \Delta \vv_{i\leftarrow j} \;=\; \vg_{i\leftarrow j} \odot \vv_j.
\end{equation}
The bond-adapter updates are then
\begin{equation}
  \Delta \mH_i \;=\; \frac{1}{|\mathcal{N}(i)|}\sum_{j \in \mathcal{N}(i)} \Delta \vh_{i\leftarrow j},
  \qquad
  \Delta \mV_i \;=\; \frac{1}{|\mathcal{N}(i)|}\sum_{j \in \mathcal{N}(i)} \Delta \vv_{i\leftarrow j},
\end{equation}
where $\mathcal{N}(i) = \{j : (i,j) \in \mathcal{E}\}$ and the scalar gate $\vg_{i\leftarrow j}$ is broadcast over the spatial axis of $\vv_j$. Because $\Delta \vh_{i\leftarrow j}$ and $\vg_{i\leftarrow j}$ are computed only from invariant quantities, they are E(3)-invariant. Because $\vv_j$ is equivariant and the gate acts only on the channel axis, $\Delta \vv_{i\leftarrow j}$ and $\Delta \mV_i$ are equivariant as well. Each atom typically has at most a few incident edges (for example, $\le 4$ covalent bonds), so the adapter has complexity $\mathcal{O}(|\mathcal{E}|) = \mathcal{O}(N)$ with a small constant factor.

In practice, the adapter can be inserted at the beginning of every block or only in the first block, or disabled altogether when no bond edge set is provided. This flexibility lets the same backbone be reused across tasks that differ in whether explicit chemical adjacency is available.

\subsection{Space Complexity Analysis}
\label{sec:complexity}

We analyze the space (memory) complexity of \model{} for a single structure with $N$ atoms, model dimension $d$, number of attention heads $H$ (head dimension $d_h = d/H$), $L$ Transformer blocks, and $|\mathcal{E}|$ edges. Following standard practice we set the feed-forward hidden dimension $d_{\text{ff}} = \Theta(d)$, so that all width-dependent terms can be expressed in $d$ alone. We separately account for \emph{parameter memory} (model weights) and \emph{activation memory} (intermediate tensors retained during inference or, more critically, during training). For batched inputs with per-sample lengths $\{L_b\}_{b=1}^{B}$, $N$ should be replaced by $\sum_b L_b$ in the activation terms.

\paragraph{Parameter memory.}
Model weights are independent of $N$ and determined only by $L$ and $d$.
\begin{itemize}
  \item \textbf{Feature initialization $\mathcal{I}$.} Scalar query/key projections contribute $\mathcal{O}(d^2)$, and the output projection from head space to $d$ channels contributes $\mathcal{O}(Hd) = \mathcal{O}(d^2)$. One-time cost is $\mathcal{O}(d^2)$.
  \item \textbf{Self-attention layers.} Each block has independent scalar $\mQ/\mK/\mU$ projections of size $\mathcal{O}(d^2)$, independent bias-free vector $\mQ/\mK/\mU$ projections of size $\mathcal{O}(d^2)$, two output projections of size $\mathcal{O}(d^2)$, and $\mathcal{O}(H)$ learnable distance scales. Per-layer cost is $\mathcal{O}(d^2)$.
  \item \textbf{Feed-forward layers.} The scalar and vector branches each maintain input and output projections of size $\mathcal{O}(d \cdot d_{\text{ff}}) = \mathcal{O}(d^2)$. Per-layer cost is $\mathcal{O}(d^2)$.
  \item \textbf{Bond adapter.} A small MLP maps $2d + d_e$ inputs to $2d$ outputs, with cost $\mathcal{O}(d^2)$ in any block where the adapter is enabled.
\end{itemize}
Summing over components, the total parameter memory is
\begin{equation}
  \underbrace{\mathcal{O}(d^2)}_{\text{init}} \;+\; \underbrace{\mathcal{O}(L d^2)}_{\text{blocks}} \;=\; \mathcal{O}(L d^2).
\end{equation}

\paragraph{Activation memory.}
\begin{itemize}
  \item \textbf{Feature initialization.} One round of distance-aware attention has the same activation profile as a single attention layer, giving $\mathcal{O}(N^2 H)$ under a naive kernel, or $\mathcal{O}(N H) = \mathcal{O}(Nd)$ with a memory-efficient attention kernel. This is a one-time cost dominated by the $L$-layer backbone below.
  \item \textbf{Self-attention.} A naive implementation materializes the $N \times N$ logit matrix per head, requiring $\mathcal{O}(N^2 H)$ memory per layer. With a memory-efficient attention kernel the full matrix is never stored, and only running softmax statistics and tile-sized buffers are maintained, reducing the per-layer cost to $\mathcal{O}(Nd)$. The distance-aware formulation appends only $9$ extra entries to queries and keys through the encoding in Equation~\eqref{eq:dist-enc}, so the effective head dimension increases from $d_h$ to $d_h + 9$ but the asymptotic class is unchanged. Crucially, no separate $N \times N$ distance-bias tensor is required.
  \item \textbf{Feature streams and feed-forward layer.} Each block maintains $\mH \in \R^{N \times d}$ and $\mV \in \R^{N \times 3 \times d}$, together occupying $\mathcal{O}(Nd)$, plus $\mathcal{O}(Nd)$ for the feed-forward intermediates. The peak memory across $L$ layers depends on the checkpointing strategy. Without checkpointing it is $\mathcal{O}(LNd)$, with $\sqrt{L}$-interval checkpointing it reduces to $\mathcal{O}(\sqrt{L}\,Nd)$ at the cost of one extra forward pass, and full recomputation stores only $\mathcal{O}(Nd)$ activations at inference time.
  \item \textbf{Bond adapter.} Message passing over a sparse bond edge set stores $\mathcal{O}(d)$ intermediate features per edge, giving $\mathcal{O}(|\mathcal{E}|d)$. Under the chemical valence assumption $|\mathcal{E}| \le cN$, this simplifies to $\mathcal{O}(Nd)$ and is dominated by the feature-stream cost.
\end{itemize}
\Cref{tab:space} summarizes these results. By (i)~using memory-efficient attention kernels and (ii)~encoding distance information through query--key augmentation rather than through an explicit bias matrix, \model{} avoids any $\mathcal{O}(N^2)$ activation term. The overall peak activation memory therefore scales \emph{linearly} in $N$ for a fixed model size, making the architecture applicable to large molecular structures.

\begin{table}[h]
\centering
\caption{Space complexity of \model{} ($d_{\text{ff}} = \Theta(d)$), where $N$ is the number of atoms, $d$ is the model dimension, $L$ is the number of Transformer blocks, and $|\mathcal{E}|$ is the number of edges with $|\mathcal{E}| = \mathcal{O}(N)$. Activation costs assume no checkpointing, with variants discussed in the text.}
\label{tab:space}
\small
\begin{tabular}{lcc}
\toprule
\textbf{Component} & \textbf{Parameters} & \textbf{Peak Activations} \\
\midrule
Initialization             & $\mathcal{O}(d^2)$   & $\mathcal{O}(Nd)$          \\
Attention (naive)          & $\mathcal{O}(Ld^2)$  & $\mathcal{O}(LN^2 d/d_h)$  \\
Attention (efficient)      & $\mathcal{O}(Ld^2)$  & $\mathcal{O}(LNd)$         \\
Feed-forward               & $\mathcal{O}(Ld^2)$  & $\mathcal{O}(LNd)$         \\
Bond adapter               & $\mathcal{O}(Ld^2)$  & $\mathcal{O}(LNd)$         \\
\midrule
\textbf{Total (efficient)} & $\mathcal{O}(Ld^2)$  & $\mathcal{O}(LNd)$         \\
\bottomrule
\end{tabular}
\end{table}

\section{Experiments}

\subsection{Datasets}
\label{sec:datasets}

We constructed our training set using a pipeline similar to that of the previously reported CPSea dataset, while extending its scope to linear peptides and adjusting several filtering thresholds.

Briefly, we use AFDB~\citep{varadi2022alphafold} domains as source structures. For each protein structure, we first compute secondary structure assignments, pLDDT scores, GRAVY scores, hydrophobicity annotations, and the residue-level C-distance matrix. We then apply sliding windows of 3--13 residues to enumerate candidate segments and retain those with average GRAVY $< 0.5$, hydrophobic residue ratio $< 0.45$, helix ratio $< 0.67$, sheet ratio $< 0.34$, minimum pLDDT $> 70$, and terminal amide C-distance within $3.5$--$15.5~\text{\AA}$ for retained segments.

The resulting candidates are further filtered using interface-based criteria derived from buried surface area (BSA), requiring total BSA $> 400~\text{\AA}^2$, relative BSA between $0.35$ and $0.85$, relative apolar BSA $< 0.75$, and limited burial of the two terminal capping side chains ($< 0.30$ each). In addition, the receptor neighborhood is required to form a connected structural graph under a residue-level connectivity cutoff of $9.0~\text{\AA}$. Candidates that pass all filters are finally clustered using a sequence-overlap threshold of $0.2$.

Starting from $8.64$ million AFDB domains, this pipeline produces approximately $100$ million candidate segments and $31$ million clusters of peptide--protein complex structures. From these, we randomly sample $100\mathrm{K}$ and $1.2\mathrm{M}$ structures for the training runs reported below.

\subsection{Memory Efficiency}

We first evaluate the memory footprint of \model{} empirically and examine whether the implementation follows the linear scaling predicted by the analysis in Table~\ref{tab:space}.

\paragraph{Setup.}
To avoid confounding the backbone comparison with variability in real complexes, we construct synthetic peptide chains of varying length at full-atom resolution.
Each chain is composed of alanine residues.
For a given residue count $n_{\mathrm{aa}}$, we enumerate all heavy atoms in the alanine template, place residues along a linear backbone with $3.8~\text{\AA}$ spacing, add small Gaussian coordinate noise, and construct both intra-residue and inter-residue covalent bonds.
We sweep $n_{\mathrm{aa}}$ over powers of two from $2$ to $1{,}024$, yielding atom counts from roughly $10$ to $5{,}000$.
For each chain length, we record the peak GPU memory allocated during a single inference forward pass without gradient computation.
\model{} and the original EPT~\citep{jiao2025equivariantpretrainedtransformerunified} backbone are evaluated under the same model dimension and number of attention heads.

\begin{figure}[t]
  \centering
  \begin{minipage}[t]{0.48\textwidth}
    \centering
    \includegraphics[width=\linewidth]{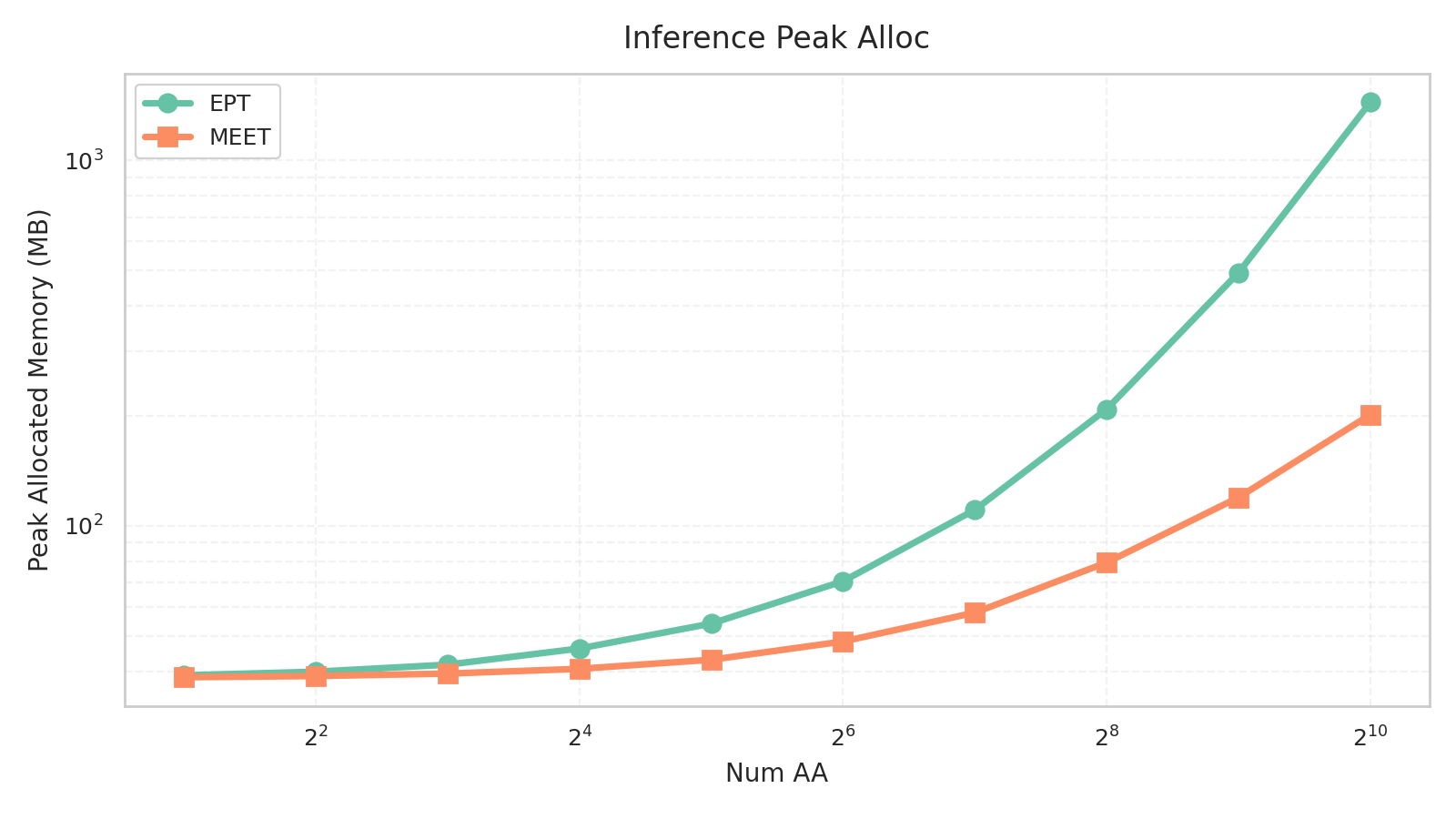}
  \end{minipage}\hfill
  \begin{minipage}[t]{0.48\textwidth}
    \centering
    \includegraphics[width=\linewidth]{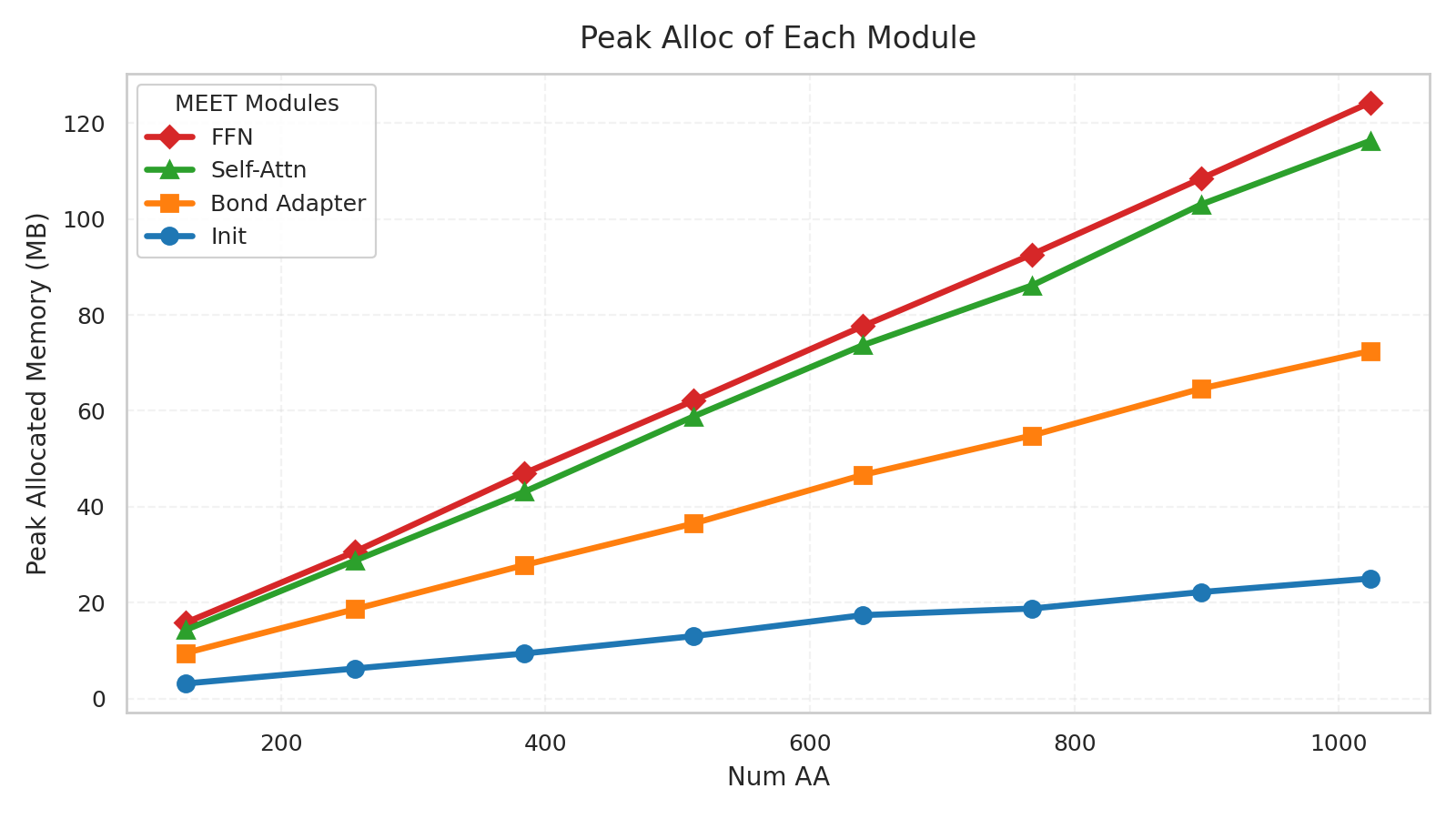}
  \end{minipage}
  \caption{\textbf{Memory efficiency of \model{}.} \emph{(Left)} Inference peak memory versus peptide length for \model{} and EPT. \model{} scales linearly, whereas EPT grows super-linearly. \emph{(Right)} Per-module memory breakdown of \model{}. All four components grow approximately linearly with sequence length.}
  \label{fig:peak_alloc}
\end{figure}

\paragraph{Overall comparison with EPT.}
Figure~\ref{fig:peak_alloc} (left) reports peak allocated memory as a function of peptide length on a log--log scale.
Across the entire range, \model{} consistently uses less memory than EPT, and the gap widens as the chain becomes longer.
At $n_{\mathrm{aa}}=1{,}024$, \model{} uses roughly $300~\mathrm{MB}$ whereas EPT exceeds $1{,}500~\mathrm{MB}$, a reduction of approximately $5\times$.
The \model{} curve remains close to linear, matching the $\mathcal{O}(LNd)$ activation scaling derived in Table~\ref{tab:space}.
By contrast, EPT follows a visibly steeper trend, consistent with the $\mathcal{O}(N^{2})$ memory cost of explicitly materializing pairwise distance-bias tensors.

\paragraph{Per-module breakdown.}
Figure~\ref{fig:peak_alloc} (right) further decomposes the memory use of \model{} into its four main components, namely FFN, self-attention, bond adapter, and initialization.
All four curves grow approximately linearly as $n_{\mathrm{aa}}$ increases from $128$ to $1{,}024$, indicating that none of the modules introduces a hidden quadratic term.
The FFN and self-attention layers account for most of the footprint, as expected from their $\mathcal{O}(Nd)$ activation tensors.
The initialization and bond-adapter modules contribute smaller but still linear overheads.

\subsection{Benchmark Results}

We next compare \model{} with existing peptide design methods on the $100\mathrm{K}$ dataset described in Section~\ref{sec:datasets}.

\paragraph{Baselines and model variants.}
We consider four recent generative baselines, PepGLAD~\citep{kong2024pepglad}, PepFlow~\citep{li2024pepflow}, UniMoMo~\citep{kong2025unimomo}, and DiffPepBuilder~\citep{wang2024diffpepbuilder}.
For our approach, we evaluate two variants, \textbf{MEET-XS} (extra-small) and \textbf{MEET-B} (base), so that the effect of backbone capacity can be assessed at a fixed data scale.

\paragraph{Evaluation metrics.}
We emphasize two metrics that capture complementary aspects of design quality.
The first is $\Delta G$, the predicted binding free energy, where lower values indicate stronger predicted binding.
The second is PoseBuster~\citep{buttenschoen2024posebusters} pass rate (PB), which measures the physical validity of generated poses and is better when higher.
We also report shape complementarity (Shape), solvation-normalized binding energy ($\Delta G / \Delta\mathrm{SASA}$), and sequence diversity (Seq.\ Div.).

\paragraph{Results.}
Table~\ref{tab:100K} presents the benchmark results.
Both MEET variants outperform all baselines on the two primary metrics.
MEET-B achieves a mean $\Delta G$ of $-27.40$ and a PoseBuster pass rate of $0.799$, compared with $-21.80$ and $0.561$ for the strongest baseline, UniMoMo.
PepGLAD and PepFlow do not pass the PoseBuster checks in this evaluation, and DiffPepBuilder reaches only $0.001$.
Even the smaller MEET-XS already surpasses every baseline, with a mean $\Delta G$ of $-25.67$ and a PB of $0.660$.
The same pattern is reflected in Shape and $\Delta G/\Delta\mathrm{SASA}$, where MEET-B obtains the best values among all methods.

The main trade-off at this scale appears in sequence diversity.
MEET-B has lower Seq.\ Div. ($0.715$) than PepGLAD ($0.939$) and UniMoMo ($0.922$), suggesting that the higher-capacity model samples a more concentrated distribution when trained on $100\mathrm{K}$ examples.
Rather than treating this as an architectural limitation, we examine in the next subsection whether the effect changes with substantially more training data.

\begin{table}[t]
\centering
\caption{Benchmark results on the $100\mathrm{K}$ training set. Best values are \textbf{bolded}. Lower $\Delta G$ and higher PB are preferred.}
\label{tab:100K}
\small
\setlength{\tabcolsep}{3.5pt}
\begin{tabular}{l cc c cc c c}
\toprule
\textbf{Method}
  & {$\Delta G$ mean}{\scriptsize$\,\downarrow$}
  & {$\Delta G$ med.}{\scriptsize$\,\downarrow$}
  & {PB}{\scriptsize$\,\uparrow$}
  & {$\frac{\Delta G}{\Delta\text{SASA}}$ mean}
  & {$\frac{\Delta G}{\Delta\text{SASA}}$ med.}
  & {Shape}
  & {Seq.\ Div.} \\
\midrule
PepGLAD        & $-18.02$ & $-17.47$ & $0.000$ & $-0.67$ & $-1.58$ & $0.566$ & $\mathbf{0.939}$ \\
PepFlow        & $-20.81$ & $-19.55$ & $0.000$ & $-2.36$ & $-2.40$ & $0.566$ & $0.798$ \\
UniMoMo        & $-21.80$ & $-22.30$ & $0.561$ & $-2.01$ & $-2.24$ & $0.633$ & $0.922$ \\
DiffPepBuilder & $\phantom{-}1.21$  & $-17.96$ & $0.001$ & $\phantom{-}0.27$  & $-1.95$ & $0.607$ & $0.843$ \\
\midrule
MEET-XS        & $-25.67$ & $-24.17$ & $0.660$ & $-2.39$ & $-2.38$ & $0.635$ & $0.838$ \\
MEET-B         & $\mathbf{-27.40}$ & $\mathbf{-25.73}$ & $\mathbf{0.799}$ & $\mathbf{-2.53}$ & $\mathbf{-2.51}$ & $\mathbf{0.651}$ & $0.715$ \\
\bottomrule
\end{tabular}
\end{table}

\subsection{Scaling to a Larger Training Set}

We then investigate how the system behaves when both the data scale and the capacity of the latent denoiser are increased.
In this experiment, the VAE is kept fixed, and the \model{} denoising network inside the LDM is scaled on the $1.2\mathrm{M}$ dataset described in Section~\ref{sec:datasets}.

\paragraph{Model configurations.}
We follow the depth and hidden-dimension conventions of DiT~\citep{peebles2023scalable}, using S, B, and L variants for the LDM denoising backbone.
We also include an XS variant with half the S-model depth.

\begin{figure}[t]
  \centering
  \begin{minipage}[t]{0.48\textwidth}
    \centering
    \includegraphics[width=\linewidth]{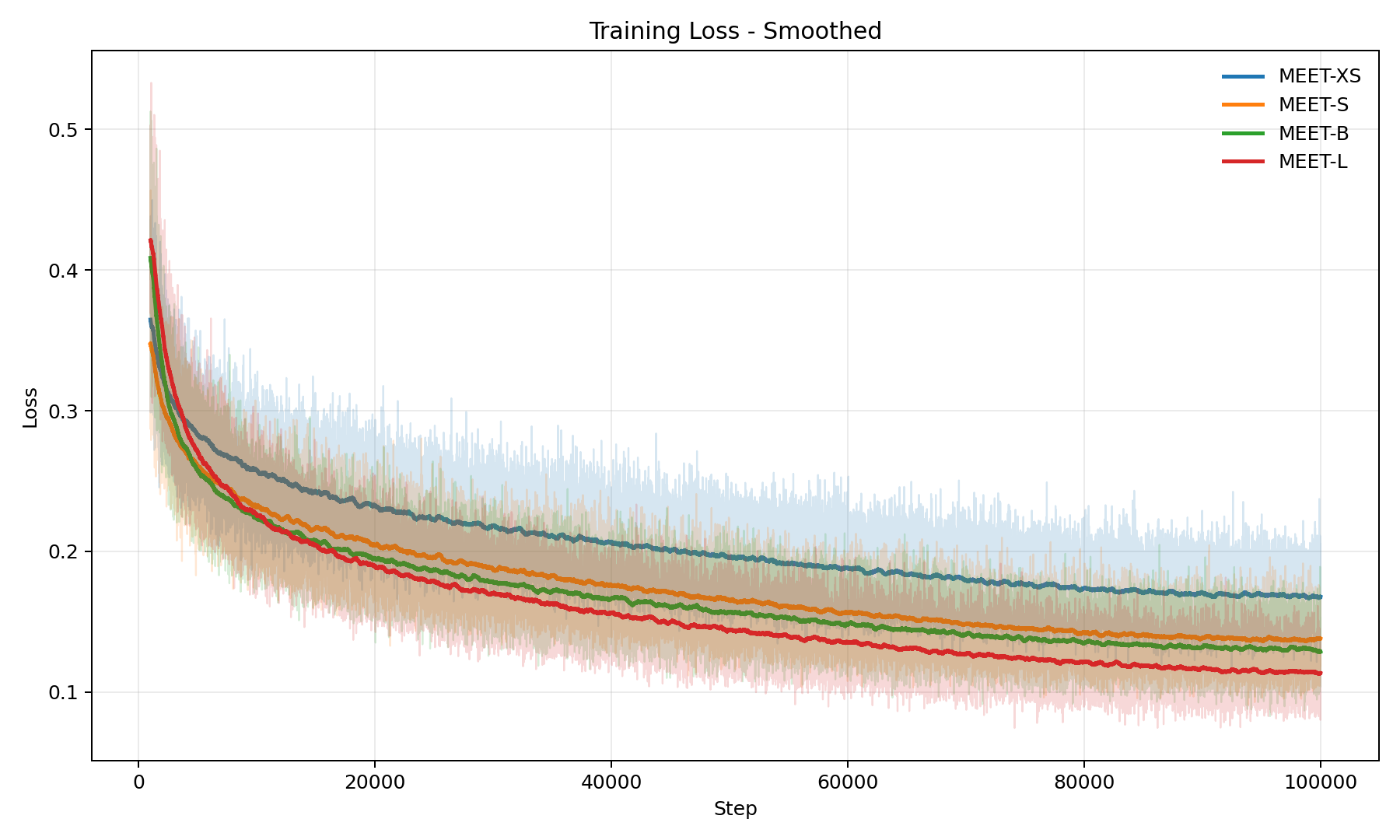}
  \end{minipage}\hfill
  \begin{minipage}[t]{0.48\textwidth}
    \centering
    \includegraphics[width=\linewidth]{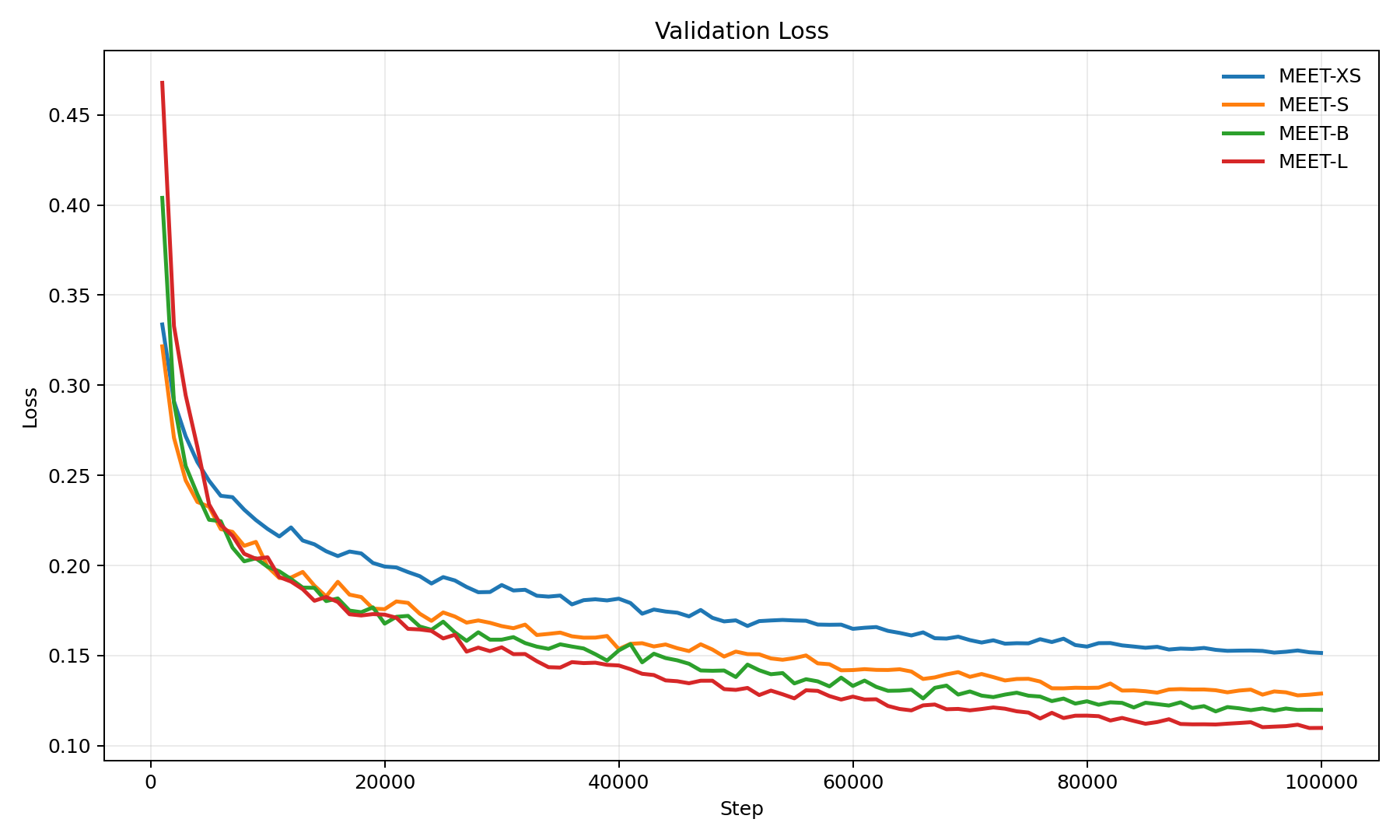}
  \end{minipage}
  \caption{Training \emph{(left)} and validation \emph{(right)} loss curves on the $1.2\mathrm{M}$ dataset. Larger LDM backbones consistently achieve lower loss, showing favorable scaling behavior.}
  \label{fig:loss_curves}
\end{figure}

\paragraph{Training dynamics.}
Figure~\ref{fig:loss_curves} shows training and validation loss curves over $100\mathrm{K}$ optimization steps.
Larger backbones consistently reach lower loss, and the ordering MEET-L\,$<$\,MEET-B\,$<$\,MEET-S\,$<$\,MEET-XS is maintained on both splits throughout training.
This trend indicates that the memory-efficient design of \model{} does not prevent the denoiser from exploiting additional capacity.
Instead, the backbone remains effective as the LDM is scaled to larger models.

\paragraph{Generation quality.}
Table~\ref{tab:1250k} reports downstream generation metrics for all four model sizes.
As capacity increases from XS to L, mean $\Delta G$ improves from $-26.26$ to $-28.22$ and PB increases from $0.703$ to $0.732$.
Shape complementarity and solvation efficiency follow the same trend.
The effect of data scale is also clear, as MEET-XS trained on $1.2\mathrm{M}$ already matches or exceeds the $100\mathrm{K}$ MEET-B setting across most metrics.

Importantly, sequence diversity improves substantially at the larger data scale.
MEET-L retains a Seq.\ Div. of $0.899$, compared with $0.715$ for MEET-B on $100\mathrm{K}$.
This suggests that the reduced diversity observed in Table~\ref{tab:100K} is primarily a data-scale effect rather than an inherent consequence of the \model{} backbone.

\begin{table}[t]
\centering
\caption{Generation quality on the $1.2\mathrm{M}$ training set across LDM backbone scales. Best results are \textbf{bolded}.}
\label{tab:1250k}
\small
\setlength{\tabcolsep}{3.5pt}
\begin{tabular}{l cc c cc c c}
\toprule
\textbf{Model}
  & {$\Delta G$ mean}{\scriptsize$\,\downarrow$}
  & {$\Delta G$ med.}{\scriptsize$\,\downarrow$}
  & {PB}{\scriptsize$\,\uparrow$}
  & {$\frac{\Delta G}{\Delta\text{SASA}}$ mean}
  & {$\frac{\Delta G}{\Delta\text{SASA}}$ med.}
  & {Shape}
  & {Seq.\ Div.} \\
\midrule
MEET-XS & $-26.26$ & $-25.06$ & $0.703$ & $-2.39$ & $-2.38$ & $0.640$ & $\mathbf{0.927}$ \\
MEET-S  & $-27.63$ & $-26.04$ & $0.727$ & $-2.52$ & $-2.51$ & $0.651$ & $0.918$ \\
MEET-B  & $-27.61$ & $-26.44$ & $0.729$ & $-2.55$ & $-2.54$ & $0.657$ & $0.915$ \\
MEET-L  & $\mathbf{-28.22}$ & $\mathbf{-27.04}$ & $\mathbf{0.732}$ & $\mathbf{-2.61}$ & $\mathbf{-2.59}$ & $\mathbf{0.659}$ & $0.899$ \\
\bottomrule
\end{tabular}
\end{table}

Taken together, these results show that the memory-efficient architecture of \model{} enables practical scaling along both the model-capacity and data axes, and that this scaling translates into consistent improvements in peptide design quality.

\section{Conclusion}

We introduced \model{}, a memory-efficient E(3)-equivariant Transformer backbone for full-atom peptide design.
Its main design principle is to express geometric computation in forms compatible with memory-efficient attention, including distance-aware query--key augmentation, global vector initialization, and sparse bond adaptation.
This removes the quadratic activation bottleneck of dense geometric attention while preserving coupled invariant and equivariant feature streams.

Integrated into a VAE and latent diffusion framework, \model{} improves both efficiency and generation quality.
The backbone shows linear memory scaling in peptide length, outperforms prior peptide design methods on binding affinity and physical validity, and supports systematic scaling of the latent denoiser on larger training data.
These results indicate that backbone efficiency is not only an implementation concern, but a key enabler for scaling full-atom generative models.

\bibliographystyle{plainnat}
\bibliography{references}

@article{wang2022therapeutic,
  title={Therapeutic peptides: current applications and future directions},
  author={Wang, Lei and Wang, Nanxi and Zhang, Wenping and Cheng, Xurui and Yan, Zhibin and Shao, Gang and Wang, Xi and Wang, Rui and Fu, Caiyun},
  journal={Signal transduction and targeted therapy},
  volume={7},
  number={1},
  pages={48},
  year={2022},
  publisher={Nature Publishing Group UK London}
}

@article{thomas2018tensor,
  title={Tensor field networks: Rotation-and translation-equivariant neural networks for 3d point clouds},
  author={Thomas, Nathaniel and Smidt, Tess and Kearnes, Steven and Yang, Lusann and Li, Li and Kohlhoff, Kai and Riley, Patrick},
  journal={arXiv preprint arXiv:1802.08219},
  year={2018}
}

@inproceedings{
    kong2025unimomo,
    title={UniMoMo: Unified Generative Modeling of 3D Molecules for De Novo Binder Design},
    author={Kong, Xiangzhe and Zhang, Zishen and Zhang, Ziting and Jiao, Rui and Ma, Jianzhu and Liu, Kai and Huang, Wenbing and Liu, Yang},
    booktitle={Forty-second International Conference on Machine Learning},
    year={2025}
}

@article{kong2024pepglad,
  title={Full-atom peptide design with geometric latent diffusion},
  author={Kong, Xiangzhe and Jia, Yinjun and Huang, Wenbing and Liu, Yang},
  journal={Advances in Neural Information Processing Systems},
  volume={37},
  pages={74808--74839},
  year={2025}
}

@inproceedings{li2024pepflow,
  title={Full-Atom Peptide Design based on Multi-modal Flow Matching},
  author={Li, Jiahan and Cheng, Chaoran and Wu, Zuofan and Guo, Ruihan and Luo, Shitong and Ren, Zhizhou and Peng, Jian and Ma, Jianzhu},
  booktitle={International Conference on Machine Learning},
  pages={27615--27640},
  year={2024},
  organization={PMLR}
}

@article{wang2024diffpepbuilder,
  title={Target-specific de novo peptide binder design with diffpepbuilder},
  author={Wang, Fanhao and Wang, Yuzhe and Feng, Laiyi and Zhang, Changsheng and Lai, Luhua},
  journal={Journal of Chemical Information and Modeling},
  volume={64},
  number={24},
  pages={9135--9149},
  year={2024},
  publisher={ACS Publications}
}

@article{varadi2022alphafold,
  title={AlphaFold Protein Structure Database: massively expanding the structural coverage of protein-sequence space with high-accuracy models},
  author={Varadi, Mihaly and Anyango, Stephen and Deshpande, Mandar and Nair, Sreenath and Natassia, Cindy and Yordanova, Galabina and Yuan, David and Stroe, Oana and Wood, Gemma and Laydon, Agata and others},
  journal={Nucleic acids research},
  volume={50},
  number={D1},
  pages={D439--D444},
  year={2022},
  publisher={Oxford University Press}
}

@article{buttenschoen2024posebusters,
  title={PoseBusters: AI-based docking methods fail to generate physically valid poses or generalise to novel sequences},
  author={Buttenschoen, Martin and Morris, Garrett M and Deane, Charlotte M},
  journal={Chemical Science},
  volume={15},
  number={9},
  pages={3130--3139},
  year={2024},
  publisher={Royal Society of Chemistry}
}

@inproceedings{peebles2023scalable,
  title={Scalable diffusion models with transformers},
  author={Peebles, William and Xie, Saining},
  booktitle={Proceedings of the IEEE/CVF international conference on computer vision},
  pages={4195--4205},
  year={2023}
}

@article{dao2022flashattention,
  title={Flashattention: Fast and memory-efficient exact attention with io-awareness},
  author={Dao, Tri and Fu, Dan and Ermon, Stefano and Rudra, Atri and R{\'e}, Christopher},
  journal={Advances in neural information processing systems},
  volume={35},
  pages={16344--16359},
  year={2022}
}

@inproceedings{schutt2021equivariant,
  title={Equivariant message passing for the prediction of tensorial properties and molecular spectra},
  author={Sch{\"u}tt, Kristof and Unke, Oliver and Gastegger, Michael},
  booktitle={International conference on machine learning},
  pages={9377--9388},
  year={2021},
  organization={PMLR}
}

@inproceedings{satorras2021en,
  title={E (n) equivariant graph neural networks},
  author={Satorras, V{\i}ctor Garcia and Hoogeboom, Emiel and Welling, Max},
  booktitle={International conference on machine learning},
  pages={9323--9332},
  year={2021},
  organization={PMLR}
}

@article{zhang2019root,
  title={Root mean square layer normalization},
  author={Zhang, Biao and Sennrich, Rico},
  journal={Advances in neural information processing systems},
  volume={32},
  year={2019}
}

@article{schutt2018schnet,
  title={Schnet--a deep learning architecture for molecules and materials},
  author={Sch{\"u}tt, Kristof T and Sauceda, Huziel E and Kindermans, P-J and Tkatchenko, Alexandre and M{\"u}ller, K-R},
  journal={The Journal of chemical physics},
  volume={148},
  number={24},
  year={2018},
  publisher={AIP Publishing}
}

@inproceedings{
tholke2022torchmdnet,
title={Equivariant Transformers for Neural Network based Molecular Potentials},
author={Philipp Th{\"o}lke and Gianni De Fabritiis},
booktitle={International Conference on Learning Representations},
year={2022},
url={https://openreview.net/forum?id=zNHzqZ9wrRB}
}

@inproceedings{
    liao2023equiformer,
    title={Equiformer: Equivariant Graph Attention Transformer for 3D Atomistic Graphs},
    author={Yi-Lun Liao and Tess Smidt},
    booktitle={International Conference on Learning Representations},
    year={2023},
    url={https://openreview.net/forum?id=KwmPfARgOTD}
}

@article{shazeer2020glu,
  title={GLU Variants Improve Transformer},
  author={Shazeer, Noam},
  journal={arXiv preprint arXiv:2002.05202},
  year={2020}
}

@inproceedings{rombach2022high,
  title={High-resolution image synthesis with latent diffusion models},
  author={Rombach, Robin and Blattmann, Andreas and Lorenz, Dominik and Esser, Patrick and Ommer, Bj{\"o}rn},
  booktitle={Proceedings of the IEEE/CVF conference on computer vision and pattern recognition},
  pages={10684--10695},
  year={2022}
}

@article{ho2020denoisingdiffusionprobabilisticmodels,
  title={Denoising diffusion probabilistic models},
  author={Ho, Jonathan and Jain, Ajay and Abbeel, Pieter},
  journal={Advances in neural information processing systems},
  volume={33},
  pages={6840--6851},
  year={2020}
}

@inproceedings{nichol2021improveddenoisingdiffusionprobabilistic,
  title={Improved denoising diffusion probabilistic models},
  author={Nichol, Alexander Quinn and Dhariwal, Prafulla},
  booktitle={International conference on machine learning},
  pages={8162--8171},
  year={2021},
  organization={PMLR}
}

@inproceedings{
  song2021scorebasedgenerativemodelingstochastic,
  title={Score-Based Generative Modeling through Stochastic Differential Equations},
  author={Yang Song and Jascha Sohl-Dickstein and Diederik P Kingma and Abhishek Kumar and Stefano Ermon and Ben Poole},
  booktitle={International Conference on Learning Representations},
  year={2021},
  url={https://openreview.net/forum?id=PxTIG12RRHS}
}

@inproceedings{
lipman2023flowmatchinggenerativemodeling,
title={Flow Matching for Generative Modeling},
author={Yaron Lipman and Ricky T. Q. Chen and Heli Ben-Hamu and Maximilian Nickel and Matthew Le},
booktitle={The Eleventh International Conference on Learning Representations },
year={2023},
url={https://openreview.net/forum?id=PqvMRDCJT9t}
}

@article{jiao2025equivariantpretrainedtransformerunified,
  title={An equivariant pretrained transformer for unified 3D molecular representation learning},
  author={Jiao, Rui and Kong, Xiangzhe and Zhang, Li and Yu, Ziyang and Ren, Fangyuan and Tan, Wenjuan and Huang, Wenbing and Liu, Yang},
  journal={Nature Communications},
  year={2026},
  publisher={Nature Publishing Group UK London}
}

@inproceedings{deng2021vectorneuronsgeneralframework,
  title={Vector neurons: A general framework for so (3)-equivariant networks},
  author={Deng, Congyue and Litany, Or and Duan, Yueqi and Poulenard, Adrien and Tagliasacchi, Andrea and Guibas, Leonidas J},
  booktitle={Proceedings of the IEEE/CVF international conference on computer vision},
  pages={12200--12209},
  year={2021}
}

@article{Chaudhury_2010,
  title={PyRosetta: a script-based interface for implementing molecular modeling algorithms using Rosetta},
  author={Chaudhury, Sidhartha and Lyskov, Sergey and Gray, Jeffrey J},
  journal={Bioinformatics},
  volume={26},
  number={5},
  pages={689--691},
  year={2010},
  publisher={Oxford University Press}
}

@inproceedings{
wu2026flashbias,
title={FlashBias: Fast Computation of Attention with Bias},
author={Haixu Wu and Minghao Guo and Yuezhou Ma and Yuanxu Sun and Jianmin Wang and Wojciech Matusik and Mingsheng Long},
booktitle={The Thirty-ninth Annual Conference on Neural Information Processing Systems},
year={2026},
url={https://openreview.net/forum?id=7L4NvUtZY3}
}

\clearpage
\beginappendix
\section{Introduction of Latent Generative Framework}
\label{app:latent-framework}

\setcounter{table}{0}
\renewcommand{\thetable}{\Alph{section}.\arabic{table}}

We instantiate \model{} in a two-stage latent generative framework for target-specific full-atom peptide design.
The overall design follows the motivation of latent generative modeling in UniMoMo~\citep{kong2025unimomo}, where detailed atomistic structures are first compressed into a compact latent representation and conditional generation is then performed in this lower-dimensional space.
This separation is useful for peptide design because the atomic system contains many local constraints, while the global design decision is naturally organized at the residue or block level.

\paragraph{Atom-level VAE.}
For a peptide--protein complex, let $\mX \in \R^{N \times 3}$ denote atom coordinates, let $A_i$ denote atom types, and let $S_b$ denote the block type of residue or fragment $b$.
The VAE maps the full-atom complex into block-level latent variables
\begin{equation}
  \mZ = (\mZ_H,\mZ_X), \qquad
  \mZ_H \in \R^{M \times d_z}, \qquad
  \mZ_X \in \R^{M \times 3},
\end{equation}
where $M$ is the number of blocks.
The invariant latent $\mZ_H$ captures block identity and local chemical context, whereas the equivariant latent $\mZ_X$ records block-level geometry.
The encoder first computes atom-level scalar and vector features with \model{}, then aggregates atom features within each block to parameterize an approximate posterior
\begin{equation}
  q_{\phi}(\mZ \mid \mX,A,S)
  =
  \prod_{b=1}^{M}
  q_{\phi}(\mZ_{H,b}\mid \mX,A,S)\,
  q_{\phi}(\mZ_{X,b}\mid \mX,A,S).
\end{equation}
We regularize the invariant latent toward a standard Gaussian prior and the coordinate latent toward a Gaussian centered at the corresponding block center $\vr_b$,
\begin{equation}
  p(\mZ)
  =
  \prod_{b=1}^{M}
  \mathcal{N}(\mZ_{H,b};\vzero,\mI)\,
  \mathcal{N}(\mZ_{X,b};\vr_b,\mI).
\end{equation}
This prior encourages each block latent point to remain close to the full-atom geometry it represents, while still allowing the latent diffusion model to later model global peptide placement.
In our framework, we do not introduce separate learned block or chain embeddings.
The block abstraction is instead induced by atom-level encoding, block-wise pooling, position information, and the latent variables themselves.

The decoder factorizes reconstruction into sequence prediction and coordinate generation.
The sequence decoder predicts block types from the latent point cloud,
\begin{equation}
  p_{\xi}(S\mid \mZ) = \prod_{b=1}^{M} p_{\xi}(S_b\mid \mZ_H,\mZ_X),
\end{equation}
and is trained with cross entropy.
Given the decoded or ground-truth block types, the structure decoder reconstructs atom coordinates under a bond graph specified by the target topology, peptide chain adjacency, predicted block identities, and terminal capping rules, avoiding a separate bond-generation distribution.

Coordinate reconstruction is trained as a continuous flow-matching problem~\citep{lipman2023flowmatchinggenerativemodeling}.
For each atom, an initial coordinate $\mX_{\mathrm{prior}}$ is sampled around the corresponding block latent coordinate, and a time $t \sim \mathcal{U}(0,1)$ is drawn.
We define the interpolation
\begin{equation}
  \mX_t = \mX_{\mathrm{prior}} + (1-t)(\mX - \mX_{\mathrm{prior}}),
  \qquad
  \vu^{\star} = \mX - \mX_{\mathrm{prior}},
\end{equation}
so that $t=1$ corresponds to the prior state and $t=0$ corresponds to the data structure.
The structure decoder predicts a vector field $\vu_{\xi}(\mX_t,t,\mZ,S)$ and is trained to match $\vu^{\star}$.
The full VAE objective can be written as
\begin{align}
  \mathcal{L}_{\mathrm{VAE}}
  &=
  \lambda_{\mathrm{seq}}\,
  \mathrm{CE}\!\left(S,\hat S\right)
  + \lambda_{\mathrm{coord}}^{\mathrm{lig}}\,
  \bigl\|\mM_{\mathrm{lig}}\odot(\vu_{\xi}-\vu^{\star})\bigr\|_2^2
  + \lambda_{\mathrm{coord}}^{\mathrm{poc}}\,
  \bigl\|\mM_{\mathrm{poc}}\odot(\vu_{\xi}-\vu^{\star})\bigr\|_2^2 \notag \\
  &\quad
  + \lambda_{\mathrm{dist}}\,\mathcal{L}_{\mathrm{dist}}
  + \lambda_H\,
  D_{\mathrm{KL}}\!\left(q_{\phi}(\mZ_H\mid \mX,A,S)\,\|\,p(\mZ_H)\right)
  + \lambda_X\,
  D_{\mathrm{KL}}\!\left(q_{\phi}(\mZ_X\mid \mX,A,S)\,\|\,p(\mZ_X)\right).
\end{align}
Here $\mM_{\mathrm{lig}}$ and $\mM_{\mathrm{poc}}$ select ligand and pocket atoms, and $\mathcal{L}_{\mathrm{dist}}$ preserves local geometric distances.
During VAE decoding, the coordinates are initialized from $\mX_{\mathrm{prior}}$ and updated along a decreasing time grid $1=t_K>\cdots>t_0=0$ by
\begin{equation}
  \mX_{t_{k-1}}
  =
  \mX_{t_k}
  +
  (t_k-t_{k-1})\,
  \vu_{\xi}(\mX_{t_k},t_k,\mZ,\hat S),
\end{equation}
which transports atoms from the latent prior toward a full-atom structure.

\paragraph{Latent diffusion model.}
After VAE training, the autoencoder is frozen and a conditional latent diffusion model is trained on the VAE latents.
Let $\mC$ denote the target-pocket context encoded by the frozen VAE, and let $\mZ_0=(\mZ_{H,0},\mZ_{X,0})$ denote the clean peptide latents.
The coordinate latent is centered by the pocket center and normalized by a fixed coordinate scale before diffusion training.

Both $\mZ_H$ and $\mZ_X$ are modeled with continuous-time cosine diffusion paths inspired by DDPM formulations~\citep{ho2020denoisingdiffusionprobabilisticmodels,nichol2021improveddenoisingdiffusionprobabilistic}.
For each latent field $r\in\{H,X\}$, the forward process samples
\begin{equation}
  \mZ_{r,t}
  =
  \eta(t)\mZ_{r,0}
  +
  \sigma(t)\vepsilon_r,
  \qquad
  t\sim\mathcal{U}(0,1),
  \qquad
  \vepsilon_r\sim\mathcal{N}(\vzero,\mI),
\end{equation}
where $\eta(t)$ and $\sigma(t)$ are the cosine schedule coefficients.
The denoising network $\vepsilon_{\theta}$ is a \model{} backbone conditioned on $\mC$ and trained to predict the injected noise.
The LDM objective is
\begin{equation}
  \mathcal{L}_{\mathrm{LDM}}
  =
  \E_{t,\mZ_0,\vepsilon_H,\vepsilon_X}
  \left[
  \left\|
  \vepsilon_{\theta}^{H}(\mZ_{H,t},\mZ_{X,t},t,\mC)-\vepsilon_H
  \right\|_2^2
  +
  \left\|
  \vepsilon_{\theta}^{X}(\mZ_{H,t},\mZ_{X,t},t,\mC)-\vepsilon_X
  \right\|_2^2
  \right].
\end{equation}

At inference time, the target pocket is first encoded into $\mC$.
Peptide latents are initialized from Gaussian noise at $t=1$, and the reverse trajectory is solved with the probability-flow ODE sampler~\citep{song2021scorebasedgenerativemodelingstochastic}, denoted as DiffusionODE in our experiments.
For a DDPM probability path, the corresponding forward SDE has drift and diffusion coefficients
\begin{equation}
  f(\mZ_{r,t},t) = -\frac{1}{2}\beta(t)\mZ_{r,t},
  \qquad
  g(t) = \sqrt{\beta(t)}.
\end{equation}
Since the denoiser predicts the noise, the score is estimated as
\begin{equation}
  \vs_{\theta}^{r}(\mZ_{H,t},\mZ_{X,t},t,\mC)
  =
  -\frac{\vepsilon_{\theta}^{r}(\mZ_{H,t},\mZ_{X,t},t,\mC)}{\sigma(t)},
  \qquad r\in\{H,X\}.
\end{equation}
The probability-flow ODE is then
\begin{equation}
  \frac{\mathrm{d}\mZ_{r,t}}{\mathrm{d}t}
  =
  f(\mZ_{r,t},t)
  -
  \frac{1}{2}g(t)^2
  \vs_{\theta}^{r}(\mZ_{H,t},\mZ_{X,t},t,\mC),
  \qquad r\in\{H,X\}.
\end{equation}
Using a decreasing time grid $1=t_K>\cdots>t_0=0$, the sampler applies an Euler update
\begin{equation}
  \mZ_{r,t_{k-1}}
  =
  \mZ_{r,t_k}
  +
  (t_{k-1}-t_k)
  \left[
  f(\mZ_{r,t_k},t_k)
  -
  \frac{1}{2}g(t_k)^2
  \vs_{\theta}^{r}(\mZ_{H,t_k},\mZ_{X,t_k},t_k,\mC)
  \right].
\end{equation}
On the final step, the solver returns the corresponding clean-latent prediction
\begin{equation}
  \hat{\mZ}_{r,0}
  =
  \frac{\mZ_{r,t}-\sigma(t)\vepsilon_{\theta}^{r}(\mZ_{H,t},\mZ_{X,t},t,\mC)}
       {\eta(t)}.
\end{equation}
This procedure produces peptide block latents at $t=0$.
The frozen VAE decoder then predicts the peptide sequence, constructs the bond graph from the decoded blocks and peptide topology, initializes atom coordinates around $\mZ_X$, and applies the coordinate flow decoder to generate the final full-atom peptide--protein complex.
In this way, \model{} is used in the encoder, the VAE decoders, and the latent denoiser, making the scalability of the backbone central to the efficiency of the entire pipeline.

\section{Evaluation Metrics}
\label{app:metrics}

The benchmark tables report five evaluation metrics, including predicted binding free energy $\Delta G$, PoseBuster pass rate (PB), shape complementarity (Shape), solvation-normalized binding energy $\Delta G/\Delta\mathrm{SASA}$, and sequence diversity (Seq.\ Div.).
These metrics are chosen to jointly assess binding affinity, physical validity, interface packing, energetic efficiency, and sample diversity.

\noindent{\bfseries\boldmath $\Delta G$.}
This metric measures the predicted binding free energy of the generated peptide against the target pocket.
For each generated complex, the structure is first relaxed under coordinate constraints, and the peptide--target interface is then scored with a PyRosetta interface energy function~\citep{Chaudhury_2010}.
Lower $\Delta G$ indicates a more favorable predicted binding interaction.
We report both the mean and the median across generated samples, since the mean reflects overall sample quality while the median is less sensitive to a small number of very strong or very weak designs.

\noindent\textbf{PB.}
This metric measures the fraction of generated poses that pass PoseBuster~\citep{buttenschoen2024posebusters}.
PoseBuster evaluates whether a generated peptide pose is chemically and geometrically plausible, including molecular parsing, bond geometry, internal clashes, flatness constraints, and intermolecular clashes with the target pocket.
A sample is counted as valid only when all selected checks are passed.
Higher PB therefore indicates that a method produces fewer physically implausible complexes.

\noindent\textbf{Shape.}
This metric denotes the PyRosetta shape-complementarity score of the peptide--target interface after the same constrained relaxation used for interface scoring~\citep{Chaudhury_2010}.
It measures how well the molecular surfaces of the generated peptide and target pocket pack against each other.
Higher Shape values indicate tighter and more geometrically complementary interfaces, independent of whether the peptide sequence matches the reference binder for that target.

\noindent{\bfseries\boldmath $\Delta G/\Delta\mathrm{SASA}$.}
This metric normalizes the predicted binding free energy by the buried solvent-accessible surface area at the interface.
This metric distinguishes designs that obtain favorable energy through efficient local interactions from those that rely mainly on forming a larger buried interface.
Lower values indicate stronger predicted binding per unit buried area.
As with $\Delta G$, we report both the mean and the median across generated samples for each method.

\noindent\textbf{Seq.\ Div.}
This metric measures the diversity of generated peptide sequences for the same target.
For each target pocket, we compare all pairs of generated sequences and compute their pairwise dissimilarity as one minus their position-wise amino-acid recovery.
The score is averaged within each target and then averaged across targets.
Higher Seq.\ Div. indicates that the model can propose a broader set of peptide sequences rather than repeatedly sampling near-identical binders.

\section{Hyperparameter Settings}
\label{app:hyperparameters}

We list only the hyperparameters that determine model capacity, latent dimensionality, objective balance, and optimization across the reported runs.

\begin{table}[t]
\centering
\caption{Key hyperparameters of the VAE.}
\label{tab:app-vae-hparams}
\small
\setlength{\tabcolsep}{5pt}
\renewcommand{\arraystretch}{1.12}
\begin{tabular}{>{\raggedright\arraybackslash}p{0.28\linewidth} c >{\raggedright\arraybackslash}p{0.44\linewidth}}
\toprule
\textbf{Name} & \textbf{Value} & \textbf{Description} \\
\midrule
\multicolumn{3}{l}{\textbf{Encoder}} \\
Backbone & \model{} & Full-atom encoder with bond-aware layers. \\
Depth & $6$ & Number of encoder layers. \\
Hidden size & $128$ & Scalar and vector channel dimension. \\
Attention heads & $8$ & Number of attention heads. \\
Latent size & $8$ & Dimension of the invariant block latent state. \\
\midrule
\multicolumn{3}{l}{\textbf{Sequence Decoder}} \\
Backbone & \model{} & Decoder for peptide block-type prediction. \\
Depth & $3$ & Number of decoder layers. \\
Hidden size & $128$ & Hidden dimension shared with the encoder. \\
Attention heads & $8$ & Number of attention heads. \\
\midrule
\multicolumn{3}{l}{\textbf{Structure Decoder}} \\
Backbone & \model{} & Full-atom coordinate decoder with specified bonds. \\
Depth & $6$ & Number of decoder layers. \\
Hidden size & $128$ & Hidden dimension shared with the encoder. \\
Attention heads & $8$ & Number of attention heads. \\
Coordinate prior std. & $1.0$ & Standard deviation of the coordinate prior. \\
Decode steps & $10$ & Number of coordinate-flow decoding iterations. \\
\midrule
\multicolumn{3}{l}{\textbf{Training Regime}} \\
Training steps & $100{,}000$ & Number of optimization steps. \\
Batch size & $1024$ & Training batch size. \\
Learning rate & $1.0\times10^{-3}$ & Initial learning rate with cosine decay. \\
Optimizer & AdamW & Weight decay $1.0\times10^{-5}$. \\
$\lambda_H$, $\lambda_X$ & $0.6$, $0.8$ & Weights for latent KL terms. \\
$\lambda_{\mathrm{seq}}$ & $1.0$ & Weight for block-type cross entropy. \\
$\lambda_{\mathrm{coord}}^{\mathrm{lig}}$ & $1.0$ & Weight for peptide-atom vector-field matching. \\
$\lambda_{\mathrm{coord}}^{\mathrm{poc}}$ & $1.0$ & Weight for pocket-atom vector-field matching. \\
$\lambda_{\mathrm{dist}}$ & $0.5$ & Weight for preserving local geometry. \\
\bottomrule
\end{tabular}
\end{table}

\begin{table}[t]
\centering
\caption{Key hyperparameters of the latent diffusion models.}
\label{tab:app-ldm-hparams}
\footnotesize
\setlength{\tabcolsep}{3pt}
\renewcommand{\arraystretch}{1.12}
\begin{tabular*}{\linewidth}{@{\extracolsep{\fill}}l l c c c c c @{\hspace{1.6em}} l l@{}}
\toprule
\multicolumn{7}{c}{\textbf{Backbone scale}} & \multicolumn{2}{c}{\textbf{Common training and sampling}} \\
\cmidrule(lr){1-7}\cmidrule(l){8-9}
\textbf{Data} & \textbf{Model} & \textbf{L} & \textbf{d} & \textbf{Heads} & \textbf{Steps} & \textbf{LR}
& \textbf{Name} & \textbf{Value} \\
\midrule
$100\mathrm{K}$ & MEET-XS & $6$  & $384$  & $6$  & $100{,}000$ & $1.0\times10^{-3}$ & Loss weights & $1.0$, $1.0$ \\
$100\mathrm{K}$ & MEET-B  & $12$ & $512$  & $8$  & $100{,}000$ & $3.0\times10^{-4}$ & Optimizer & AdamW \\
$1.2\mathrm{M}$ & MEET-XS & $6$  & $384$  & $6$  & $100{,}000$ & $1.0\times10^{-3}$ & Weight decay & $1.0\times10^{-5}$ \\
$1.2\mathrm{M}$ & MEET-S  & $12$ & $384$  & $6$  & $100{,}000$ & $5.0\times10^{-4}$ & LR schedule & Cosine \\
$1.2\mathrm{M}$ & MEET-B  & $12$ & $768$  & $12$ & $100{,}000$ & $3.0\times10^{-4}$ & Sampler & Prob.-flow ODE \\
$1.2\mathrm{M}$ & MEET-L  & $24$ & $1024$ & $16$ & $200{,}000$ & $1.0\times10^{-4}$ & Sampling steps & $100$ \\
\bottomrule
\end{tabular*}
\end{table}

For all LDM variants, the VAE is frozen during LDM training and the learning rate follows cosine decay.
In our runs, we find that larger LDM backbones required smaller learning rates for stable and effective optimization, with XS, S, B and L using $1.0\times10^{-3}$, $5.0\times10^{-4}$, $3.0\times10^{-4}$, and $1.0\times10^{-4}$, respectively.

\end{document}